\documentclass[11pt,a4paper,twocolumn]{article}

\usepackage[utf8]{inputenc}
\usepackage[T1]{fontenc}
\usepackage{amsmath,amssymb,amsfonts}
\usepackage{graphicx}
\usepackage{natbib}
\usepackage{hyperref}
\usepackage{xcolor}
\usepackage{booktabs}
\usepackage{xspace}
\usepackage{subcaption}
\usepackage{stfloats}
\usepackage{tikz}
\usetikzlibrary{shapes.geometric, arrows, calc, positioning, fit, backgrounds}

\usepackage[activate={true,nocompatibility},final,tracking=true,kerning=true,spacing=true]{microtype}
\microtypesetup{expansion=false}

\usepackage[margin=1in]{geometry}
\usepackage{authblk}

\hypersetup{
	colorlinks=true,
	linkcolor=blue,
	filecolor=magenta,
	urlcolor=cyan,
	citecolor=blue,
	pdftitle={An Early Warning Model for Forced Displacement},
	pdfauthor={Geraldine Henningsen},
	pdfsubject={Anticipatory Action, Forced Displacement, Risk Modeling},
	pdfkeywords={Forced displacement, Anticipatory action, Risk modeling, Machine learning, Refugee flows},
}

\title{Gridding Forced Displacement using Semi-Supervised Learning}

\author[1]{Andrew Wells}
\author[1]{Geraldine Henningsen}
\author[2]{Brice Bolane Tchinde Kengne}
\affil[1]{Global Data Service, UNHCR, Marmorvej 51, 2100, Denmark.}
\affil[2]{Regional Bureau West and Central Africa, UNHCR, route du King Fahd Palace, 3125 Dakar, Senegal}

\date{}  

\begin{document}
	
	\maketitle
	
	\begin{abstract}
	\begin{small}	
	We present a semi-supervised approach that disaggregates refugee statistics from administrative boundaries to 0.5-degree grid cells across 25 Sub-Saharan African countries. 
	By integrating UNHCR's ProGres registration data with satellite-derived building footprints from Google Open Buildings and location coordinates from OpenStreetMap Populated Places, our label spreading algorithm creates spatially explicit refugee statistics at high granularity. 
	This methodology achieves 92.9\% average accuracy in placing over 10 million refugee observations into appropriate grid cells, enabling the identification of localized displacement patterns previously obscured in broader regional and national statistics. 
	The resulting high-resolution dataset provides a foundation for a deeper understanding of displacement drivers. 
	\end{small}  
	\end{abstract}

\section*{Introduction}
Forced displacement presents a complex humanitarian challenge that requires detailed geographical understanding to address its root causes effectively. However, current refugee statistics primarily exist at national or regional levels, with more granular data available only for specific crises situations or operational levels. 
This limitation hinders the comprehensive analyses that are needed to understand the often more localized displacement drivers. 
Without methodological innovations that reveal these nuanced patterns, humanitarian responses remain mostly reactive rather than preventive, potentially missing opportunities for targeted interventions in emerging displacement hotspots.

Previous research has attempted to address this data deficit through various approaches, including satellite imagery analysis and survey-based methods. 
While valuable, these approaches often suffer from temporal limitations or geographical restrictions that prevent the consistent application across diverse displacement contexts. 
The resulting fragmented statistics on forcibly displaced populations, 
prevent more thorough, and especially longitudinal analyses of displacement drivers and thereby of our understanding of forced displacement itself. 

This study introduces a semi-supervised learning approach that disaggregates country and regional refugee counts from 25 Sub-Saharan African countries across West, East, and Central Africa to a 0.5-degree grid cell resolution (approximately 55 × 55 kilometres). 
The methodology transforms broad statistical aggregates of displaced populations into detailed geographical insights covering 6,221 grid cells across the region of interest. 

Our approach integrates data from UNHCR's internal ProGres registration database with satellite-derived information from Google Open Buildings \citep{Google2022} and location coordinates from OpenStreetMap Populated Places \citep{HOT2022}. 
This integration allows us to overcome the limitation that only a third of UNHCR registration records contain valid data at the lowest administrative level, 
which prevents sub-national displacement statistics at a highly granular level across wider regions and areas.

Our modelling results demonstrate high accuracy in predicting refugee distribution patterns throughout East, West, and Central Africa, with an average accuracy per admin2 unit of 0.929 when placing sub-national observations into appropriate grid cells. 
This high precision enables humanitarian organizations, researchers, and policymakers to identify previously obscured displacement patterns and root causes. 

The following sections will introduce this work in more detail. 
Section two gives a brief literature review, followed by section three introducing the data in more detail. 
Sections four and five discuss the preprocessing of the data and model architecture, respectively, while section six discusses the results and section seven concludes. 

\section*{Literature overview}

Current research on forced displacement modelling requires methodological innovations to address significant data limitations. 
\citet{Hoffmann2023} emphasize the necessity for standardized methodologies to predict displacement movements accurately as global displacement levels continue to rise. 
Organizations actively engaged in displacement prediction and response, such as UNHCR, the Internal Displacement Monitoring Centre (IDMC), and the Danish Refugee Council (DRC), regularly produce statistical analyses and forecasts of displacement trends. These analyses, however, predominantly operate at national scales, limiting their practical application for localized humanitarian response.

Some initiatives have attempted to address this geographical limitation through sub-national modelling approaches. 
DRC's AHEAD model represents one such effort, providing sub-national-level predictions rather than solely national aggregates. 
This model's application remains geographically constrained, focusing exclusively on the Liptako-Gourma border region between Mali, Niger, and Burkina Faso, with additional coverage of South Sudan and Somalia. 
The geographical specificity, while allowing for detailed analysis within these regions, prevents broader application of the model across diverse displacement contexts throughout Africa or beyond.

Recent advances in population disaggregation methodologies offer promising approaches that can be adapted to displacement modelling. 
\citet{Stevens2015} pioneered a semi-automated dasymetric modelling approach incorporating detailed census and ancillary data within a flexible Random Forest estimation technique. 
Their methodology utilizes remotely sensed and geospatial data to generate gridded population density predictions at approximately 100-meter resolution. This prediction layer subsequently serves as a weighting surface for the dasymetric redistribution of country-level census counts. 
Building upon this work, \citet{Wardrop2018} address the challenge of producing spatially disaggregated local-scale population estimates in contexts where census data may be outdated, inaccurate, or missing relevant areas. 
Their work demonstrates how recent technological advances in satellite imagery, geopositioning tools, statistical methods, and processing power enable population distribution estimation at fine spatial resolutions across entire countries, even without comprehensive census data.

Alternative data sources have further expanded the possibilities for population disaggregation in data-scarce environments. 
\citet{Patel2017} explore the potential of geo-located social media data, specifically Twitter posts, as a covariate in population mapping. 
Their research shows that including the geographical distribution of Twitter posts as an additional data source in random forest models significantly improves population mapping accuracy throughout Indonesia. 
This approach highlights the value of non-traditional data sources for enhancing spatial demographic understanding. 
Similarly, \citet{Sorichetta2015} develop methodologies for creating high-resolution gridded population distributions by harmonizing multiple data sources. Their work through the WorldPop Project has produced open-access archives of fine-resolution gridded population data covering wide geographical areas.

Building footprint data derived from satellite imagery represents another promising direction for spatial disaggregation of population statistics. 
\citet{Reed2018} compare multiple built settlement datasets for improving population disaggregation, demonstrating how variations in settlement identification methodologies affect resulting population distribution estimates. 
While their study predates Google Open Buildings, their findings on the importance of the quality of building footprints  for accurate population disaggregation remains relevant to our work. 
\citet{Tiecke2017} further advance this approach through Facebook's population mapping initiative, which disaggregates population counts using building footprints identified through computer vision combined with limited ground-truth training data. 
Their methodology demonstrates particularly promising results for regions where the collection of traditional census data faces substantial challenges.

Our methodology builds upon these various strands of research by adapting spatial disaggregation techniques specifically to data on the origin of refugees. 
While previous aggregation approaches have primarily focused on the general population distribution or, in displacement contexts, on destination locations, our work addresses the gap in understanding the geographical origins of forcibly displaced populations. 
By combining semi-supervised learning with building footprint data and refugee registration records, we extend existing methodologies and produce statistics on the origin of refugees at the 0.5° grid cell resolution across East, Central, and West Africa. 

\section*{Data}
Our methodological framework uses three complementary data sources that enable spatial disaggregation of refugee origin patterns: UNHCR's ProGres registry, Google Open Buildings \citep{Google2022}, 
and OpenStreetMap Populated Places \citep{HOT2022}. 
These three datasets provide the foundation for translating administrative-level refugee counts into grid cell estimates across our study region.

\begin{figure*}[h!]
	\includegraphics[width=0.9\textwidth]{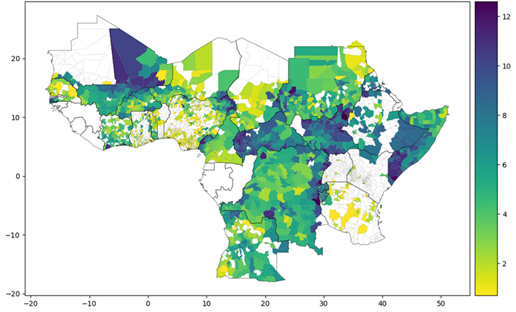}
	\caption{Total logged forced displacement outflows at the Admin 2 level for the years 2000 - 2022}
	\label{fig:one}
\end{figure*}

UNHCR's ProGres registry represents this study's primary source of refugee displacement information. 
Developed in 2002 as a comprehensive case management tool, ProGres is a centralized information repository documenting approximately 18 million registered refugees across more than 130 countries. 
Each registry entry contains detailed individual-level data, including asylum country, arrival date, demographic characteristics (age, gender, ethnic group), and, most important for our use case, hierarchical place of origin information spanning administrative levels from the country (admin0) to village (admin3). 
This hierarchical geographic structure provides essential spatial reference points for our disaggregation methodology, though with substantial variation in completeness across administrative levels. 
The origin location, given through the information at the admin1, admin2 and admin3 level, of each ProGres observation is used in developing this method. 

As shown in Figure~\ref{fig:one}, the study focuses on 25 countries across East, Central, and West Africa, selected for their data availability and their experience with significant recent displacement patterns. Temporal boundaries constrain the analysis of displacement events between 2000 and 2022, encompassing 10,894,618 individual registrations within the ProGres system from these focus countries.

Building footprint data from Google Open Buildings provides information on settlement patterns that inform the spatial disaggregation process. 
This dataset delivers building footprint information derived from high-resolution satellite imagery processed through deep learning models. Coverage extends across most African territories and portions of Asia, where each identified structure is accompanied by a confidence score that indicates the detection reliability. 
Our methodology utilizes the centroid point location of each building footprint to assign structures to corresponding 0.5\textdegree grid cells and to administrative districts. 
This assignment process enables the approximation of population distribution patterns within administrative units and overlapping 0.5\textdegree grid cells, thereby creating the weighting surface necessary for disaggregating refugee counts from the administrative level to the grid cell resolution.

OpenStreetMap Populated Places data complements the building footprint information by providing specific geographic coordinates for named settlements. 
This dataset offers precise point locations for towns, villages, and cities across the study region. 
These settlement coordinates serve as spatial anchors for admin3-level place names appearing in the ProGres registry. 
Through a spatial joining process, we use these data to match available admin3 entries from refugee records in ProGres with corresponding settlement locations, thereby we can place these observations within the appropriate 0.5\textdegree grid cells even when higher-level administrative boundaries span multiple grid cells.

\section*{Data processing}

The preparation of the spatial data requires several processing steps to translate building distribution patterns into the appropriate weights to disaggregate refugee counts. 
As discussed in the previous section, we use building footprints from Google Open Buildings as proxies for population distributions across our case region. Deriving proper weights from these data, requires the assignment of the building footprints to both administrative units and grid cells. 
This dual assignment of the building footprints creates the foundation for developing proportional distribution priors across the study region.

\begin{figure}[h!]
	\includegraphics[width=0.5\textwidth]{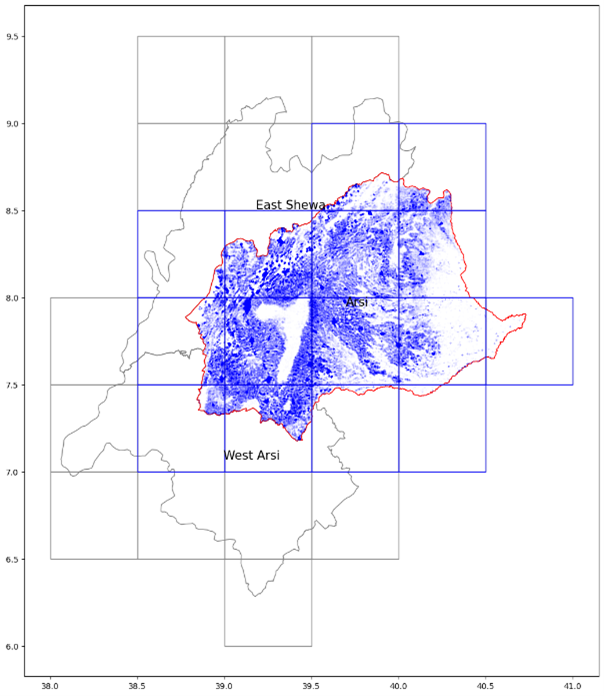}
	\caption{Overlap of 0.5\textdegree grid cells and admin2 unit (Arsi) and building footprint of admin2 unit.}
	\label{fig:two}
\end{figure}

In a first step, we determine the administrative location of the building footprints from Google Open Buildings. 
Each structure receives an administrative assignment based on its geographic coordinates, creating a dataset where individual buildings are labelled with their corresponding admin2 district. 
These same building footprints are subsequently placed within 0.5\textdegree grid cells through spatial intersection with their location data. 
The resulting dataset contains each building observation with both its admin2 designation and grid cell location.

In a second step, we then transform individual building records into proportional distributions. 
The process begins with grouping observations based on combined administrative and grid cell identifiers, generating count values for each unique spatial combination (grid cell and admin 2 unit). 
These counts are converged into a matrix where admin2 regions form the rows, grid indices for the columns, and cell values are the building counts for each admin2-grid cell combination. 
We then normalize these values to produce proportional distributions across the grid cells, where each cell value represents the percentage of an administrative unit's buildings count that is located within a specific grid cell. 
These proportions are then used as weighting factors for disaggregating refugee counts from administrative to grid resolution.  

A visual representation of these data is depicted in Figure~\ref{fig:two}. 
The image shows the Ethiopian zones (admin2) of East Shewa, West Arsi, and Arsi. 
The zone of Arsi is bordered in red, and the blue grids are the 0.5\textdegree grid cells in our data set that have at least some intersections with the Arsi administrative region. 
Hence, a displaced individual from Arsi could potentially come from any of these blue grid cells. 
However, accounting for the building distribution within Arsi, shown in blue points, offers a tool for more effectively predicting which grid cell an observation might come from. 

Correspondingly, Table~\ref{tab:griddata} offers an example of how the resulting dataset is formatted. 
Each row is a unique admin2 value and each column is a unique grid cell ID. 
The values are the proportion of buildings from a given admin2 which lie within a specific grid cell. 

\begin{table*}[htbp]
	\centering
	\begin{footnotesize}
	\begin{tabular}{l|ccccccc}
		\hline
		admin2 & grid\_10673 & grid\_10674 & grid\_10675 & grid\_10766 & grid\_10767 & grid\_10768 & grid\_10769 \\
		\hline
		East Shewa & 0 & 0 & 0.042187 & 0 & 0 & 0.092170 & 0.240284 \\
		West Arsi & 0 & 0 & 0.087890 & 0 & 0 & 0 & 0 \\
		Arsi & & 0.096668 & 0.409349 & 0.020288 & 0.005822 & 0.136971 & 0.166527 \\
		\hline
	\end{tabular}
	\caption{Data by administrative region and grid cell}
	\label{tab:griddata}
	\end{footnotesize}
\end{table*}

Cleaning the ProGres data follows a parallel workflow that is designed to maximize valid geolocation information that is contained in the database. 
First we filter observations to retain records with origin location information at minimum admin2 level. 
Administrative boundary shapefiles provide the reference framework against which we validate ProGres entries through spatial joining operations. 
This process identifies ProGres observations with valid admin2 values.

Naming inconsistencies between ProGres records and official administrative designations necessitate the implementation of approximate matching techniques. We apply a fuzzy matching algorithm employing Levenshtein similarity ratios to address minor spelling variations, typographical errors, and alternative naming conventions. The algorithm calculates the similarity between ProGres entries and standardized admin2 names using the formula:
\begin{eqnarray}
	Ratio = \left( 1 - \frac{D}{L} \right) \cdot 100 
\end{eqnarray}
where $D$ represents the Levenshtein distance \citep{Levenshtein1966} (minimum edits required to transform one string to another), and $L$ indicates the length of the longer string. 
Only matches achieving 80\% or greater similarity are accepted, balancing tolerance for minor variations against the need for precision. 

We apply a similar process to admin3 and admin4 values when admin2 matching fails, which allows us to maximize data retention while maintaining locational accuracy.

Observations that contain valid admin3 values are directly placed into 0.5° grid cells through coordinates derived from the Populated Places data. Observations from admin2 units where all buildings fall within a single grid cell similarly receive a deterministic assignment, as their spatial distribution requires no modelling. 
These observations are later reintegrated with the modelled results.

Hence, the preprocessing workflow produces three distinct datasets for each country:
\begin{enumerate}
\item Cases where admin3 information is available and which can directly be placed into grid cells.
\item Observations where all buildings in the admin2 fall within a single grid cell.
\item Observations where admin2 areas span multiple grid cells. Each observation from this data set includes:
\begin{itemize}
\item The observation’s admin2 of origin
\item A grid cell label (completed if the observation has a valid admin3 value, empty otherwise)
\item A set of columns indicating the proportion of buildings from the admin2 located in each potential grid cell
\end{itemize}
\end{enumerate}

\section{Model training}
The semi-supervised learning process employs an iterative approach that processes administrative units individually. 
Dataset segmentation begins with separation by unique admin2 IDs, analysing each administrative unit as a discrete entity. 
This segmentation ensures that the spatial context remains consistent within the modelling procedures. 

\begin{figure*}[h!]
	\includegraphics[width=\textwidth]{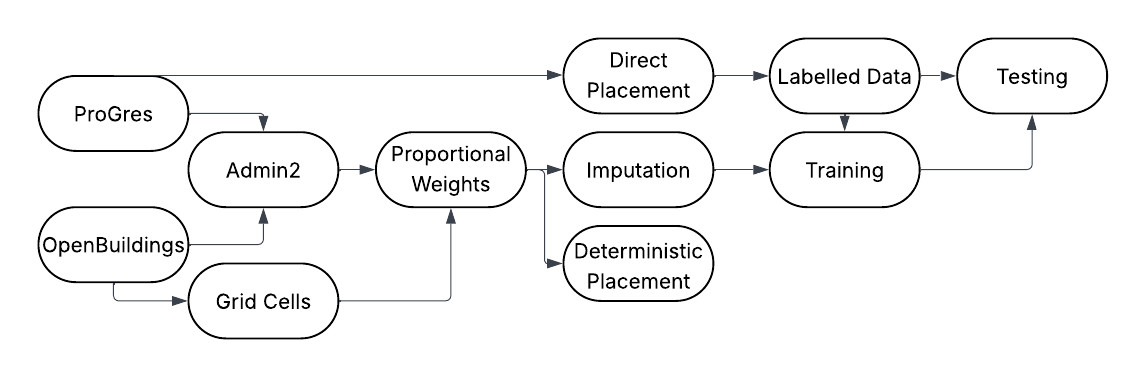}
	\caption{Modelling pipeline}
	\label{fig:five}
\end{figure*}

As outlined in Figure~\ref{fig:five}, we first identify which grid cells geographically intersect with each admin2 district. 
As described in the previous section, we then calculate building counts for these intersecting admin2-grid cell combinations, naturally resulting in non-zero building counts only for grid cells that overlap with the district. 
By retaining only these non-zero columns in our dataset, we enforce the spatial constraint that refugee observations from a specific admin2 can only be assigned to grid cells that actually intersect with that district's boundaries.

Within each admin2 unit, observations undergo categorization into two distinct sets based on spatial precision: labelled observations containing documented grid cell assignments and unlabelled observations lacking this spatial reference. 
The labelled dataset undergoes stochastic partitioning into training and validation subsets through random sampling procedures. 
For implementation of the semi-supervised learning methodology, these partitioned datasets undergo recombination to create a composite dataset incorporating both labelled training observations with their associated spatial references and unlabeled observations without grid cell affiliation.

Both categories have identical feature sets—specifically, the proportional building distribution vectors across all potential grid cells intersecting their respective admin2 boundaries. 
For instance, when an admin2 region intersects with three grid cells containing 20\%, 50\%, and 30\% of the region's building infrastructure respectively, every observation originating from this administrative unit would possess a corresponding feature vector [0.2, 0.5, 0.3]. 
This distributional representation functions as a spatial probability prior, establishing initial expectations regarding population distribution within the administrative boundaries.

The algorithmic implementation employs a label spreading methodology \citep{Zhou2003} applied to this integrated dataset. 
The algorithm constructs a similarity matrix connecting observations based on feature proximity and additional attribute characteristics (demographic variables, temporal dimensions, etc.). 
Utilizing known spatial assignments from the labelled training subset as initialization points, the algorithm executes iterative propagation of these spatial designations to adjacent unlabelled observations within the constructed similarity space.

The fundamental assumption of this approach is that systematic deviations from building-based distribution patterns observed in the labelled dataset---patterns which potentially are influenced by demographic, temporal, or other factors---can be generalized to similar unlabelled observations. 
The building distribution, hereby, provides the spatial weighting that establishes the baseline probability distribution of of origin of displaced individuals.

Through recursive iteration, the missing spatial assignments progressively converge toward equilibrium values that simultaneously satisfy the constraints imposed by the distribution of the characteristics of the labelled observations  and the prior population distribution, as proxied through the building footprint. 
Following convergence, we validate the predictive accuracy using the held-out test observations and their documented grid cell labels. 
This validation methodology tests the model's capacity to generalize learned distributional patterns to previously unobserved data points.

\begin{figure*}[htpb]
	\includegraphics[width=\textwidth]{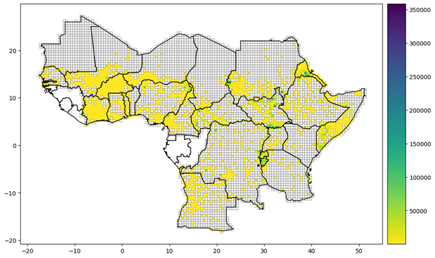}
	\caption{Gridded forced displacement counts after training totalled over 2000 to 2022.}
	\label{fig:three}
\end{figure*}

The complete disaggregation process enables the placement of over 10 million ProGres observations from 25 countries into  0.5\textdegree grid cells across the study region. 
Among the 6,221 grid cells covering the focus area, 1,785 cells (28.7\%) have recorded displacement events between 2000 and 2022. 
According to model predictions and available data, the remaining 4,436 cells (71.3\%) show zero displacements (see Figure~\ref{fig:three}.

\begin{table*}[htbp]
	\centering
	\begin{tabular}{l|cc}
		\hline
		\textit{Metric} & \textit{Semi-supervised Modeling Only} & \textit{Combined (Modeled + Deterministic)} \\
		\hline
		Accuracy & 0.845 & 0.929 \\
		F1 Score & 0.837 & 0.925 \\
		Precision & 0.842 & 0.928 \\
		Recall & 0.845 & 0.929 \\
		\hline
	\end{tabular}
	\caption{Performance metrics for grid cell placement}
	\label{tab:performance_metrics}
\end{table*}

Table~\ref{tab:performance_metrics} demonstrates that our approach shows strong predictive capabilities across multiple statistical measures. 
Figure 4 shows the distribution of the accuracy across all admin2 units in our sample. 
Although accuracy measures vary across the admin2 units, it shows that for the majority of admin2 units, high accuracy levels are reached. 
When combined with deterministically placed observations, overall performance increases substantially across all metrics. 

\begin{figure}[htpb]
	\includegraphics[width=0.5\textwidth]{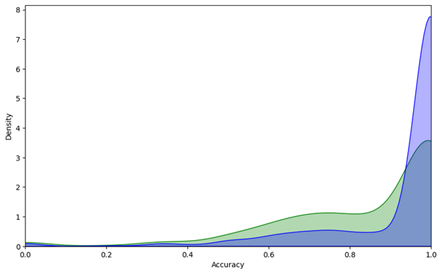}
	\caption{Kernel density plot of model accuracy across admin2 areas. Observations that underwent the semi-supervised modelling are shown in green and all observations are shown in blue.}
	\label{fig:four}
\end{figure}

The methodology encounters three limitations. 
First, prediction biases may occur and predictive uncertainty increase in areas with sparse building distribution or limited available labelled observations. 
The reliance on building footprints as population proxies potentially underrepresent displacement from sparsely built environments within administrative units despite a strong average accuracy. 

Second, observations that only contain origin information limited to admin1 fall outside the current methodology's scope, as admin1 data is too coarse for our methodology. 
The development of supplementary approaches for these coarser-resolution records continues, based on distributional patterns of admin2 observations within the same admin1 units and timeframes. 

Third, internally displaced persons (IDPs) remain uncaptured by this methodology due to its focus on refugee and asylum seeker registrations in the ProGres database. 
Internal displacement's fluid and often undocumented nature presents distinct challenges requiring specialized approaches beyond this technical report's scope.

\section*{Conclusion}
The semi-supervised learning approach developed in this study successfully places refugee observations from UNHCR's ProGres database into a standardized grid framework with high accuracy. 
Overall, our method achieves 92.9\% accuracy in assigning over 10 million refugee records to appropriate 0.5\textdegree grid cells across 25 East, Central, and West African countries. 
This spatial disaggregation transforms administrative-level refugee statistics into consistent geographic units suitable for detailed analysis of the root causes of displacement.

By using building footprint data with partial geographic labels, our approach overcomes limitations in conventional refugee statistics that typically remain aggregated at higher administrative levels. 
The resulting grid-based dataset can help to identify localized displacement factors that previously remained obscured in broader regional and national analyses.

\bibliographystyle{apalike}
\bibliography{references}  

\end{document}